\documentclass[10pt,journal,final]{IEEEtran}
\usepackage {array}
\usepackage{booktabs}
\usepackage{makecell}
\usepackage{cite}  
\usepackage{amsmath,amssymb,amsfonts}
\usepackage{textcomp}
\usepackage{xcolor}    
\usepackage{amsthm}  
\usepackage{graphicx}  
\usepackage{multirow}  
\usepackage{algpseudocode}
\usepackage[colorlinks,
linkcolor=blue,
anchorcolor=blue,
citecolor=blue]{hyperref}
\usepackage{indentfirst}
\usepackage{subfigure}

\ifCLASSINFOpdf
\else
\fi
\begin{document}

%
\title{Using Global Land Cover Product as Prompt for Cropland Mapping via Visual Foundation Model}
%
%
%

\author{Chao~Tao,~Aoran~Hu,~Rong~Xiao,~Haifeng~Li,~and~Yuze~Wang~
\thanks{Chao Tao, Aoran Hu, Rong Xiao, Haifeng Li, Yuze Wang are with the School of Geosciences and Info-Physics,  Central South University,  Changsha 410083, China (e-mail:wyz1933200059@csu.edu.cn).}
\thanks{The code in this letter are available online at https://github.com/wangyuze-csu/APT}
\thanks{Manuscript received April 19, 2020; revised August 26, 2020.}}

%
%

\markboth{Journal of \LaTeX\ Class Files,~Vol.~14, No.~8, August~2015}%
{Shell \MakeLowercase{\textit{et al.}}: Bare Demo of IEEEtran.cls for IEEE Journals}
%



\maketitle

\begin{abstract}
Data-driven deep learning methods have shown great potential in cropland mapping. However, due to multiple factors such as attributes of cropland (topography, climate, crop type) and imaging conditions (viewing angle, illumination, scale), croplands under different scenes demonstrate a great domain gap. This makes it difficult for models trained in the specific scenes to directly generalize to other scenes. A common way to handle this problem is through the “Pretrain+Fine-tuning" paradigm. Unfortunately, considering the variety of features of cropland that are affected by multiple factors, it is hardly to handle the complex domain gap between pre-trained data and target data using only sparse fine-tuned samples as general constraints. Moreover, as the number of model parameters grows, fine-tuning is no longer an easy and low-cost task. With the emergence of prompt learning via visual foundation models, the “Pretrain+Prompting" paradigm redesigns the optimization target by introducing individual prompts for each single sample. This simplifies the domain adaption from generic to specific scenes during model reasoning processes. Therefore, we introduce the “Pretrain+Prompting" paradigm to interpreting cropland scenes and design the auto-prompting (APT) method based on freely available global land cover product. It can achieve a fine-grained adaptation process from generic scenes to specialized cropland scenes without introducing additional label costs. To our best knowledge, this work pioneers the exploration of the domain adaption problems for cropland mapping under prompt learning perspectives. Our experiments using two sub-meter cropland datasets from southern and northern China demonstrated that the proposed method via visual foundation models outperforms traditional supervised learning and fine-tuning approaches in the field of remote sensing.
\end{abstract}

\begin{IEEEkeywords}
Cropland Mapping, Prompt Learning, Visual Foundation Model, Global Land Cover Product.
\end{IEEEkeywords}

%
\IEEEpeerreviewmaketitle

\section{Introduction}
%
%
%
%

\IEEEPARstart{C}{ropland} mapping is a fundamental task in agricultural resource monitoring, the accurate delineation of cropland extent has a crucial role in agricultural production management, planning, and policy formulations\cite{SAM1}. Recently, with the development of data-driven deep learning methods, they have shown strong potential in automatically extracting high-level and hierarchical features of complex scenes. And the semantic segmentation methods supported by deep learning and remote sensing data are widely applied in agriculture, urban planning, environmental monitoring, and other fields. Especially in the agricultural monitoring system\cite{SAM2, SAM3}, the intelligent farmland monitoring method via deep learning model provides a more efficient and convenient way to obtain basic information for the producers\cite{SAM4}. 

However, in the real application environment, cropland is not only affected by its own attributes\cite{SAM5} such as topography, climate, planting mode, and types of crops, but also influenced by other factors such as view angle, illumination, and scale in the imaging process\cite{SAM6}. These factors cause cropland under different spatio-temporal scenes to show strong variability in feature distribution, which leads to a complex domain gap among cropland data\cite{SAM7}. It makes the models trained in particular scenes difficult to generalize directly to other scenes. A common way to handle this challenge is through the “Pretrain+Fine-tuning” paradigm\cite{SAM8}, which freezes part of the parameters of the pre-trained model and re-trains the remains with only a small amount of labeled samples forcing the model to adapt to the target data. Although this paradigm has achieved many successful applications in a variety of remote sensing segmentation tasks\cite{SAm9}, it is still difficult to handle the domain gap between cropland scenes. The reason for this problem is that only sparse labeled samples are hard to represent complete target cropland data with diverse features, and forcing constraints on the model using this sparse information as a supervised signal may cause overfitting problem or even negative transfer. As the development of the visual foundation models\cite{SAM10}, it shows a strong ability to learn the general representation by massive data support, which gradually becomes a good choice for pre-trained models in remote sensing field. Although visual foundation models can provide more general knowledge that may suit the target data, it is still challenging to efficiently adapt general knowledge to specific cropland scenes under the influence of complex factors. Moreover, as the number of parameters for visual foundation models continues growing, the number and diversity of labeled samples required for the fine-tuning process are becoming higher, which makes it no longer an easy and low-cost task. 

With the emergence of prompt learning\cite{SAM11}, the “Pretrain+Prompting” paradigm no longer relies on sparsely labeled samples to enable the pre-trained model to adapt target data with diverse features. Instead, it transformed into the adaptation process between pre-trained data and a single target sample, which can simplify the domain gap caused by multiple factors. Specifically, in the domain adaption under the “Pretrain+Prompting” paradigm, the pre-trained model will be guided to learn the reasoning process for each sample by introducing additional prompt information. The lightweight prompt encoder will be employed while preserving the existing knowledge in the pre-trained image encoder, which simplifies the domain adapting process from generic to specific data without considering the complex multiple factors of the whole target data\cite{SAM12}. Most recently, the Segment anything model (SAM)\cite{SAM13} has emerged as a generalized framework for vision segment tasks, which shows great generalization abilities and improves efficiency by introducing various forms of prompting information, and has been widely applied in a variety of fields such as healthcare\cite{SAM14}, manufacturing\cite{SAM15}, and astronomy\cite{SAM16}. we believe that prompt learning via the visual foundation model is a potential way to address the complexity of the domain gap faced during cropland mapping. However, in more specialized remote sensing scenes with diverse features, this process still needs manual interactions\cite{SAM17} or additional labeling costs\cite{SAM18}, which makes it difficult to achieve inexpensive and efficient cropland mapping.

Therefore, instead of using sparsely labeled samples that forces the model to adapt to complex cropland scenes.We utilize the prompts to guide the model's reasoning process for each specific target sample, which can simplify the domain adaption process. Beside, during the cropland mapping for large-scale, manually collecting the prompt information for each sample  is undoubtedly costly and inefficient. To address this problem, we take the freely available global land cover (GLC) products\cite{SAM19} as the source of prompt information, which can easily obtaioned the prompts for every cropland scenes. By the auto-prompting (APT) via vision foundation model, we can extract diverse croplands without introducing any additional label costs, and provide available technical support for cropland mapping with higher spatial resolution at large scales. We conducted experiments on two sub-meter-scale cropland datasets from southern and northern China, and the result demonstrated that the proposed method via visual foundation models outperforms traditional supervised learning and fine-tuning approaches in the field of remote sensing. Moreover, we also discuss the transfer paradigm with the support of vision foundation model, and further explore the label efficiency of the auto-prompting compared with fine-tuning. The main contributions of this work are as follows:

\begin{itemize}
\item We consider the complex transfer problem caused by multiple factors such as cropland attribute and imaging conditions under prompt learning perspective, and lower the generalization difficulty from generic to specific scenes by guiding the pre-trained model’s reasoning process with remote sensing priori information. 
\item  The proposed method can automatically obtain remote sensing priori information from GLC products and guide the model’s adaption as prompt, which can further enhance the automation capability of prompt learning via the vision foundation model to realize the mapping process of diversified cropland scenes at large scale without any label cost.
\end{itemize}

\section{Principles and Methods}
\subsection{Overview of “Pretrain+Prompting” paradigm}
\begin{figure}[!t]
	\centering
	\includegraphics[width=3.5in]{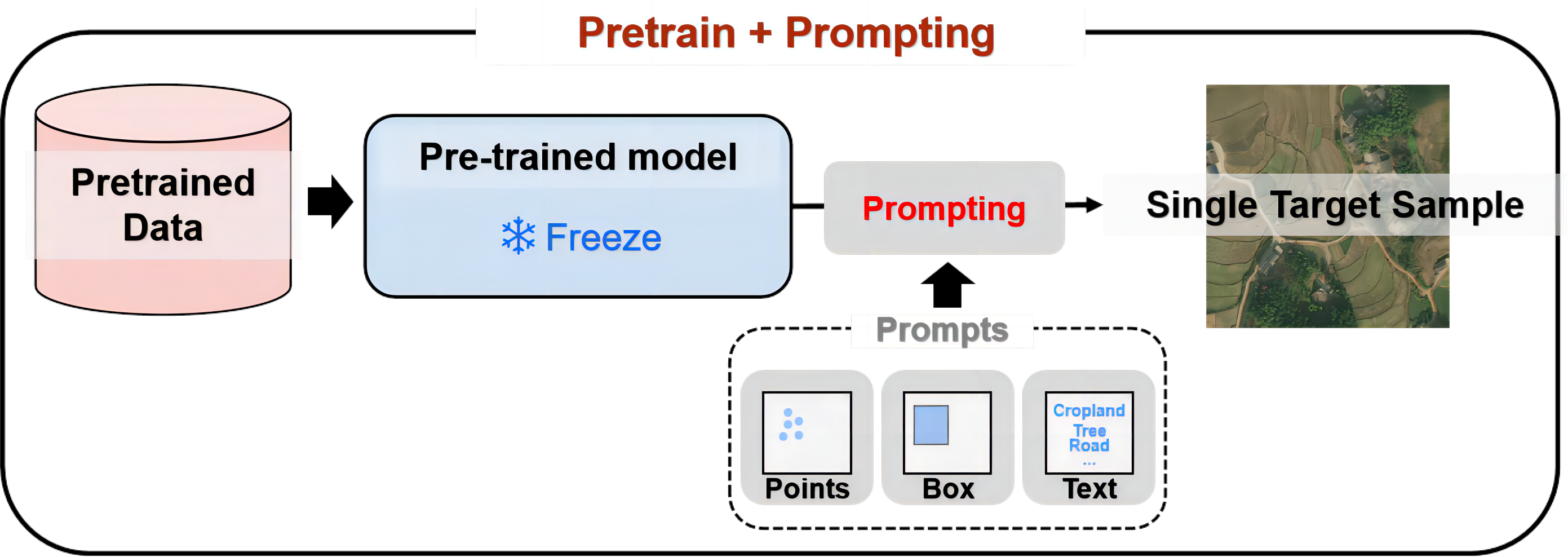}
	\caption{The overview of “Pretrain+Prompting” paradigm.}
	\label{fig1}
\vspace{-0.35cm}
\end{figure}
Under the traditional “Pretrain+Fine-tuning” paradigm, It utilizes sparsely labeled samples to guide the pretrained model to adapt the target data, which lead to the domain adaption between pretrained and target data. However, due to cost and human bias problems in the sampling process, the sparsely labeled sample sets is hard to completely represent target cropland data with diverse features. Instead of the domain adaption between data, the “Pretrain + Prompting” paradigm transforms it into the domain adaption from pretrained data to a single target sample. Specifically, It encodes flexible prompts (such as points, box or text) and images at the same time, and take the generated prompt embedding as the constrain to continuously modified the representation space under the predefined template\cite{SAM20}, which guides the model to obtain the reasoning ability for a specific scene\cite{SAM21}. In addition, compared to fine-tuning, prompting inputs additional priori information to directly optimize the model's representation of the final prediction, which allows for a near real-time adaptation process \cite{SAM12}. As shown in Figure.\ref{fig1}, after obtaining the vision foundation model through pretrained data, it guides the pretrained model to adapt single target sample $x$ by utilizing the prompt prior information $p$:
\begin{align}
	F_{target}=D_{prompt}[F_{pre},(x,p)]
\end{align}
where $F_{pre}$ and $F_{target}$ is pretrained vision fundation model and target model respectively. $D_{prompt}$ represents the domain adaption process of prompting. By this way, we can simplified the domain adaption from pretrained data to cropland target data influenced by multiple factors, which adapt pretrained model to one sample at one time. 

\subsection{The proposed workflow for cropland mapping via prompting and GLC product}
\begin{figure*}[!t]
	\centering
	\includegraphics[width=5.9in]{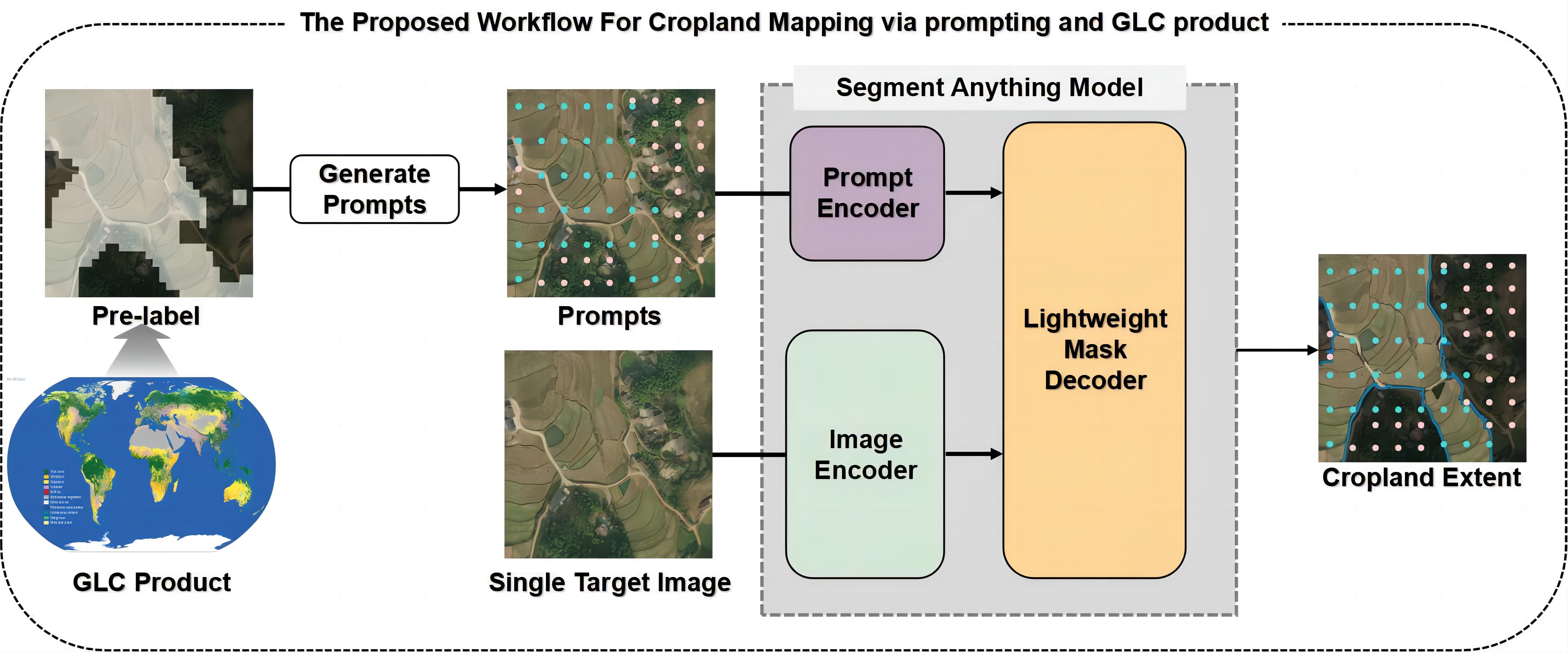}
	\caption{The proposed workflow for cropland mapping via prompting and GLC product.}
	\label{fig2}
\vspace{-0.35cm}
\end{figure*}

Although the impact of complex factors can be alleviated by adapting the model to every single target sample through prompt learning, the cost of manually collecting prompts information for each sample is undoubtedly tremendous for cropland mapping in large-scale. Beside, considering the characteristics of cropland that can be easily confused with different land cover types during the year, the collection process needs to be conducted by people with specialized knowledge, which further increase the manual cost of cropland mapping. To solve the problem, we utilize the cropland layer in the GLC product as the prompts information source of the “Pretrain+Prompting” paradigm and choose the point form to prompting the model. By  the proposed auto-prompting method (APT), we can quickly generate prompts without any human involvement and additional cost. As shown in Figure. \ref{fig2}, The proposed workflow is as follows:

\textbf{Step 1 Get Pre-label}: We obtain the cropland layer from the GLC product according to the corresponding geographic coordinate of the single target image, which generates the pre-label of cropland extent. 

\textbf{Step 2 Generate Prompts Points}: Based on the pre-label of cropland extent, we count the proportion of cropland and non-cropland area separately, and use this as a reference to generate positive and negative prompt points uniformly in the corresponding area.

\textbf{Step 3 Predict by Image and Prompts}: Input the generated prompt points and the image into the prompt encoder and image encoder respectively, and the mask decoder will efficiently map the image embedding and prompt embedding to generate the cropland possibilities.

It is worth noting that although the prompt information we employed is vague and may contain errors, the vision foundation model benefits from the pretraining process with random noise and demonstrates certain robustness, which allows the model to autonomously correct the effects of some erroneous prompt information. In addition, when the prompt information is used to guide the model to obtain the reasoning ability to understand the target scene, it is beneficial to maintain the same number and uniform distribution of positive and negative prompts to construct a more balanced and comprehensive representation space.
\begin{table*}[!h]
	\centering
		\caption{visualized examples of other traditional approaches and proposed approach.}
		\label{tab1_2}
		\begin{tabular}{cccccc}
			\includegraphics[height=3cm]{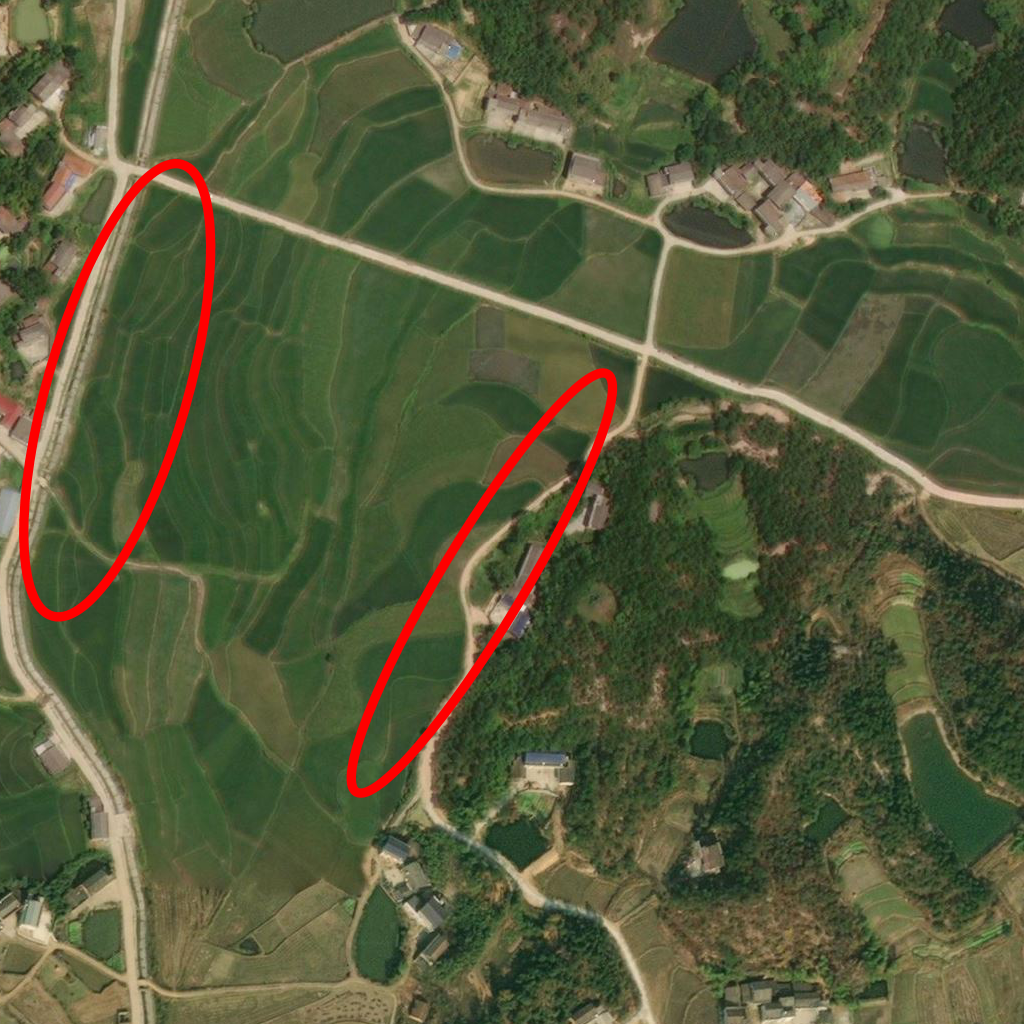}&\includegraphics[height=3cm]{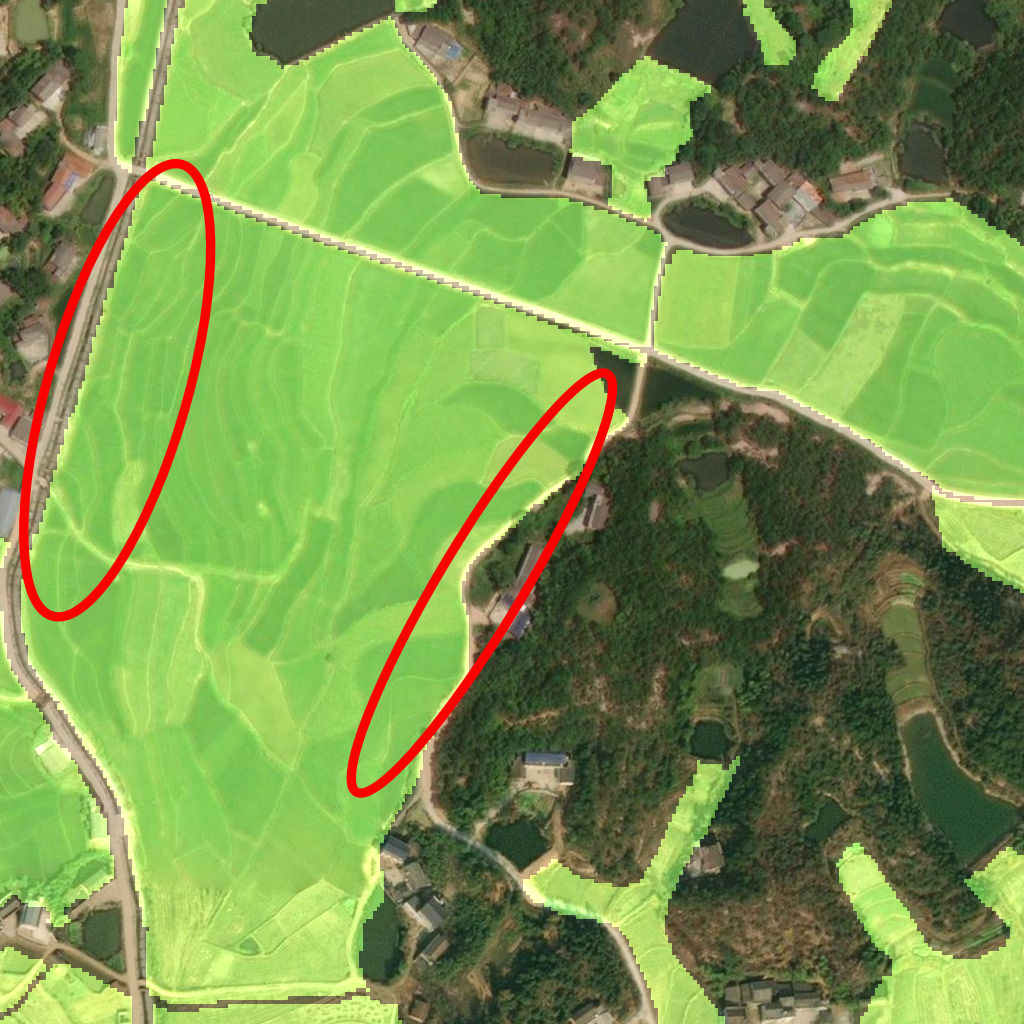}&\includegraphics[height=3cm]{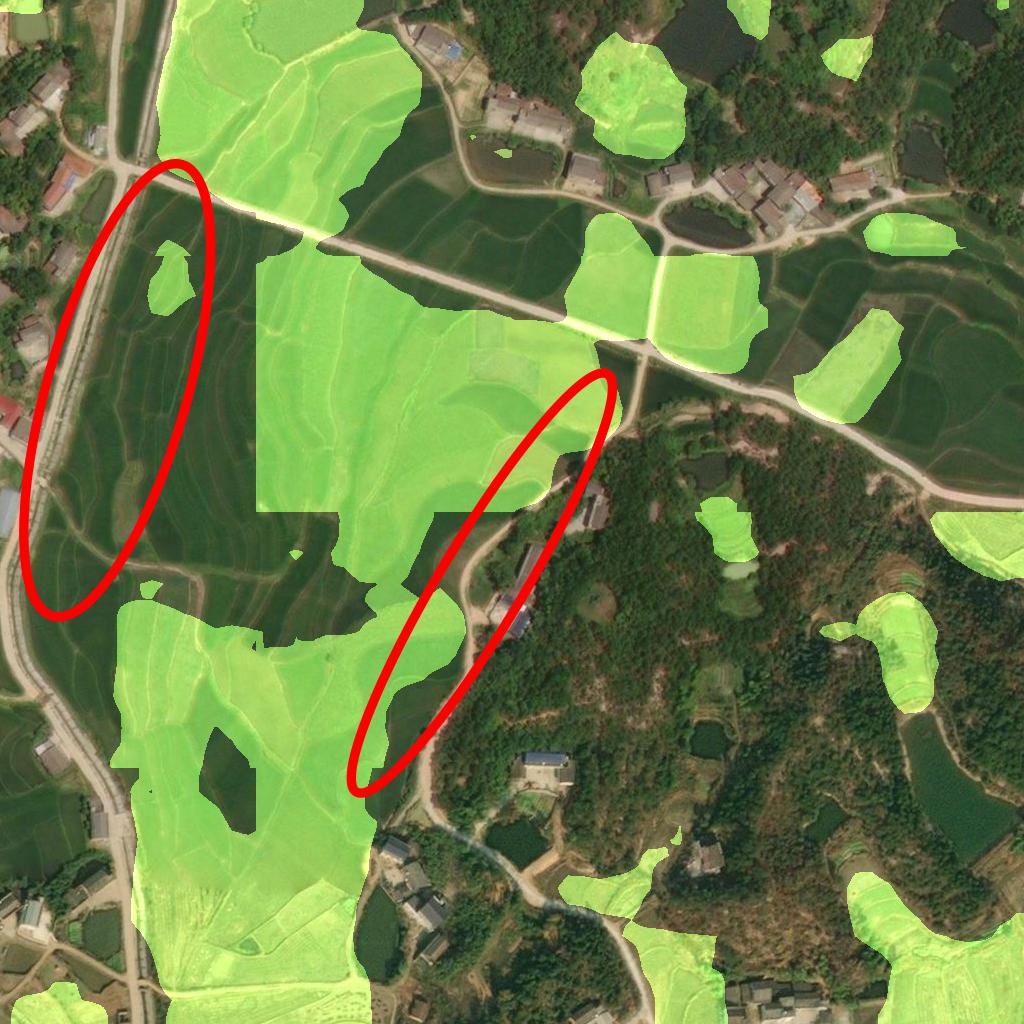}&\includegraphics[height=3cm]{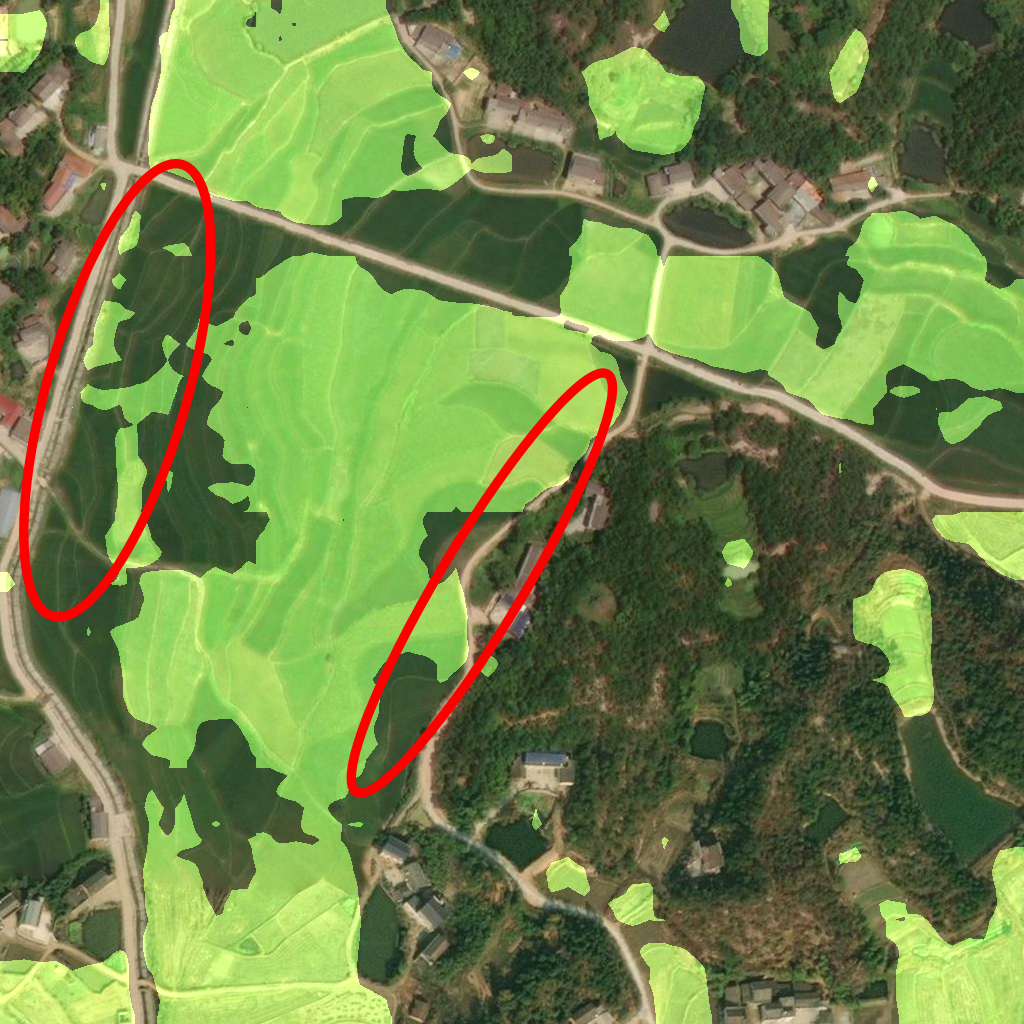}&\includegraphics[height=3cm]{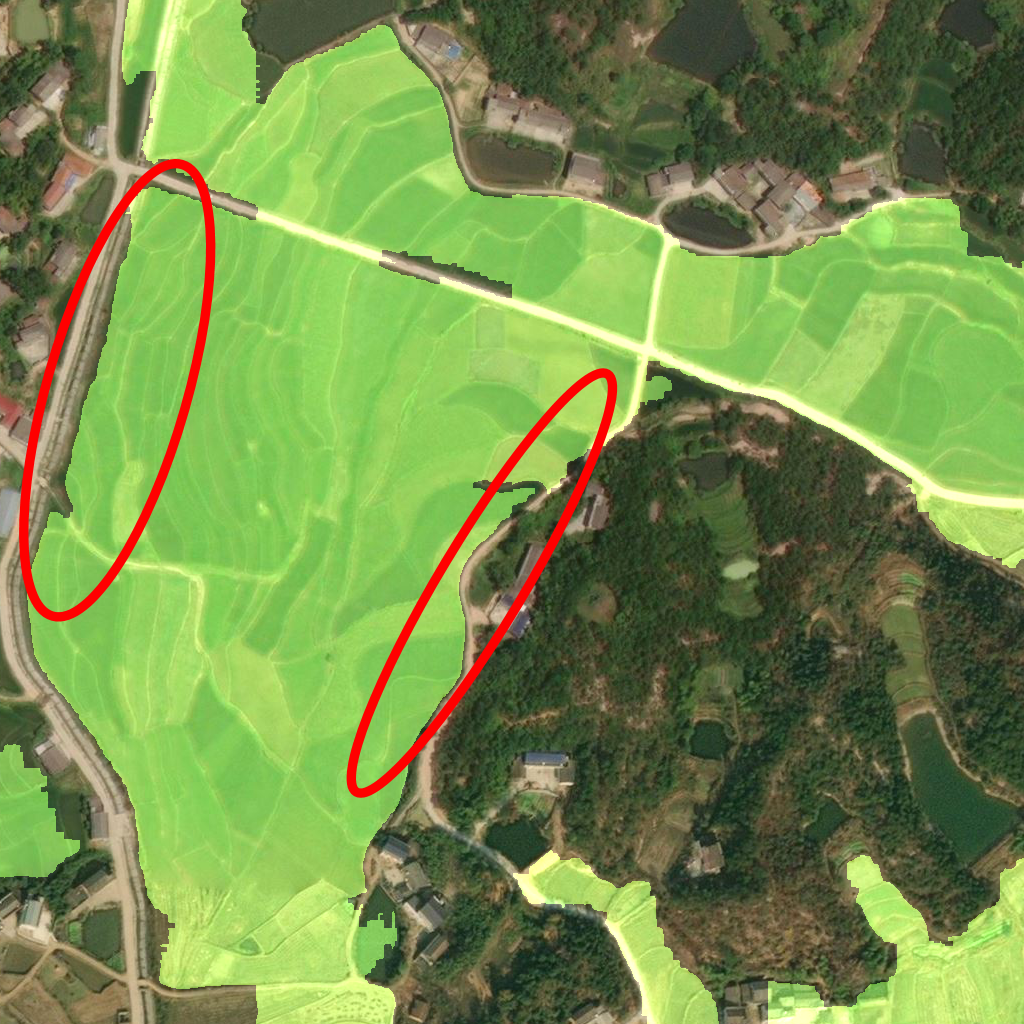} \\
			\includegraphics[height=3cm]{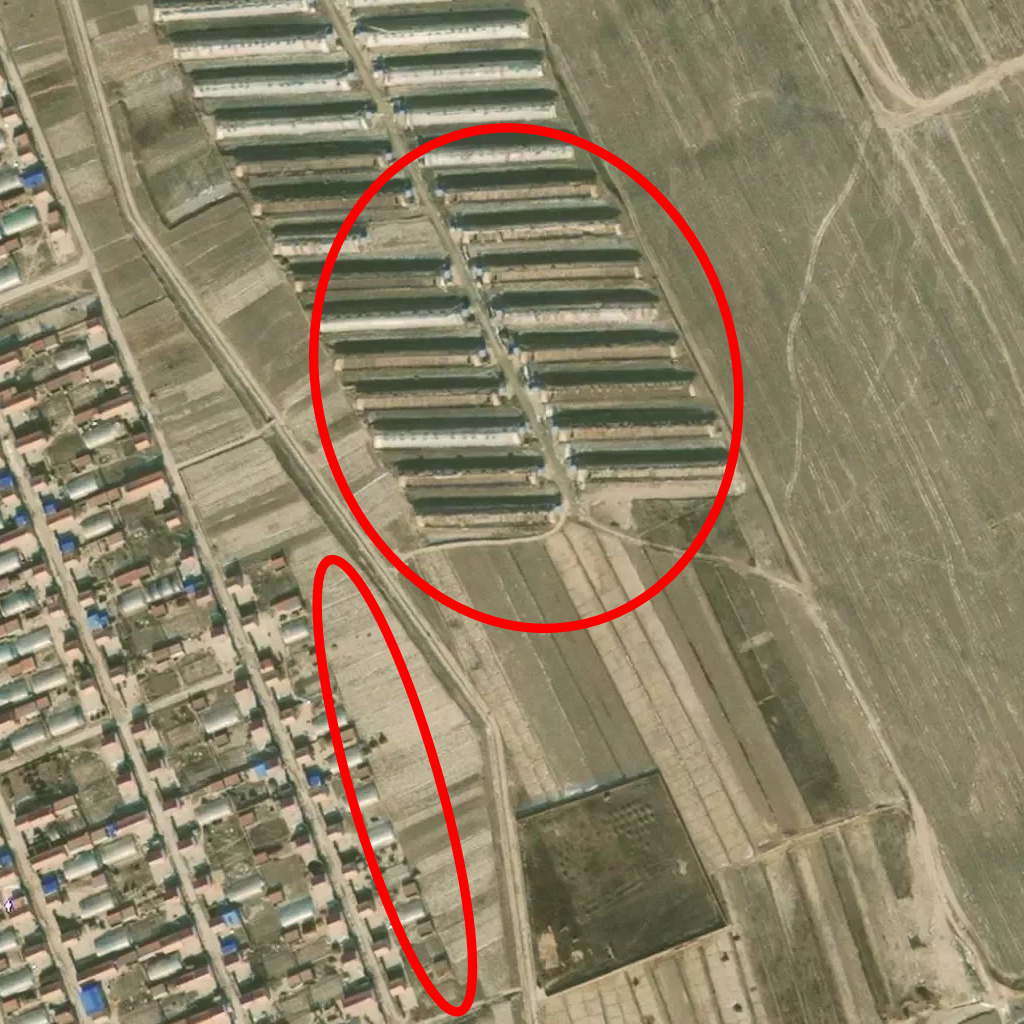}&\includegraphics[height=3cm]{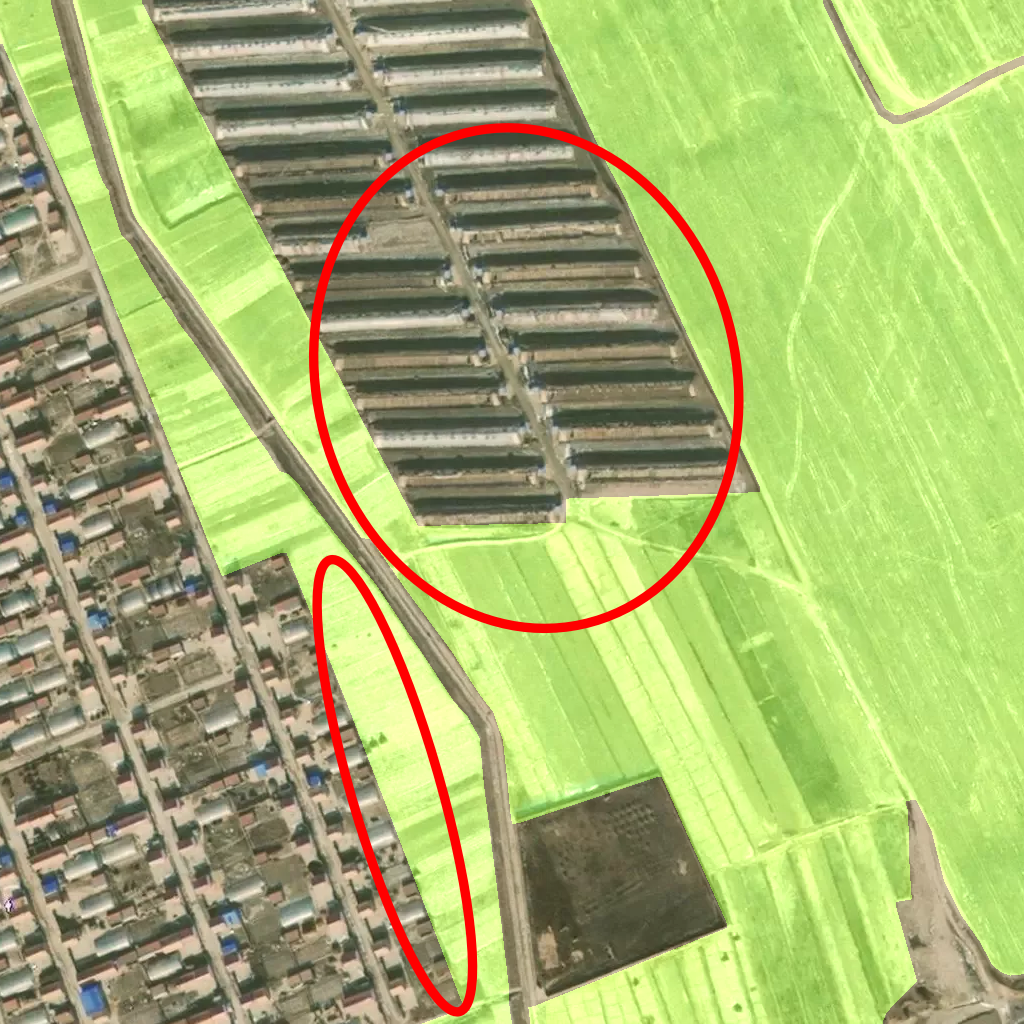}&\includegraphics[height=3cm]{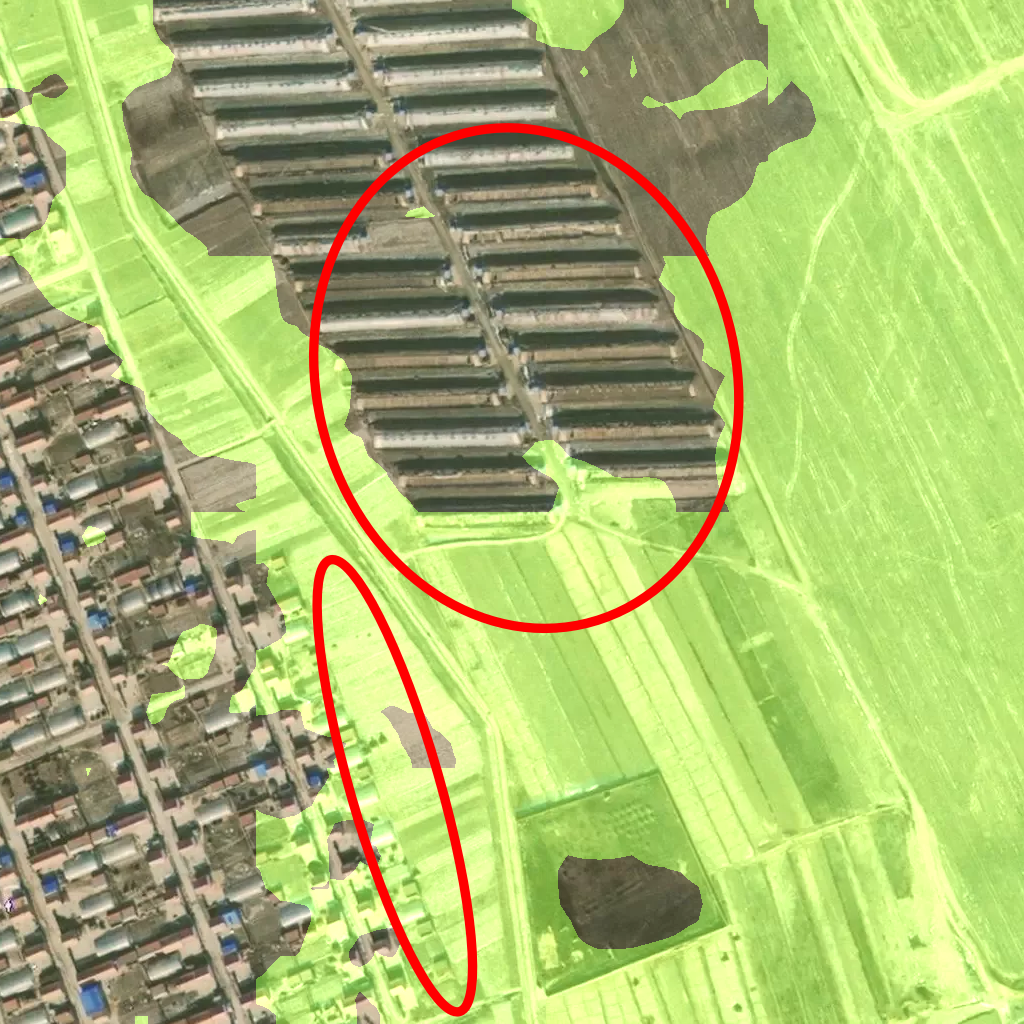}&\includegraphics[height=3cm]{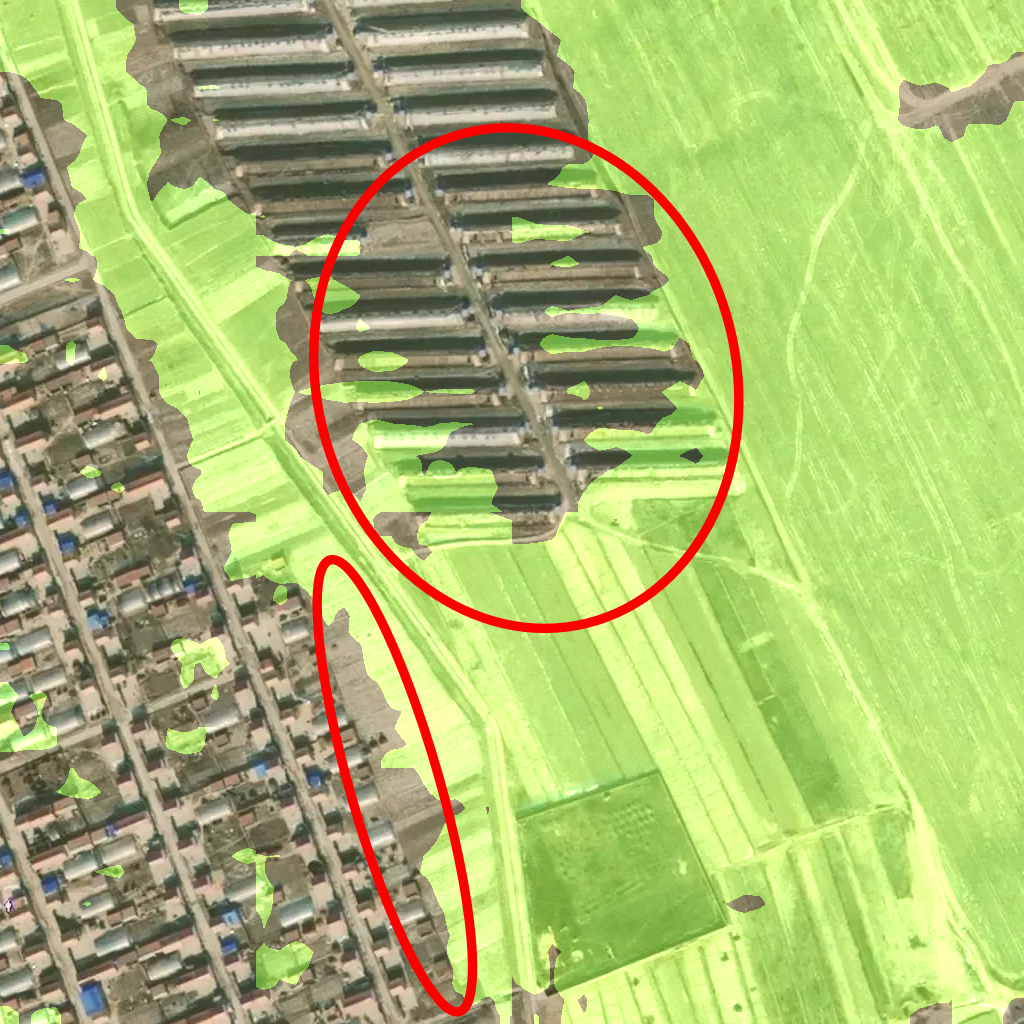}&\includegraphics[height=3cm]{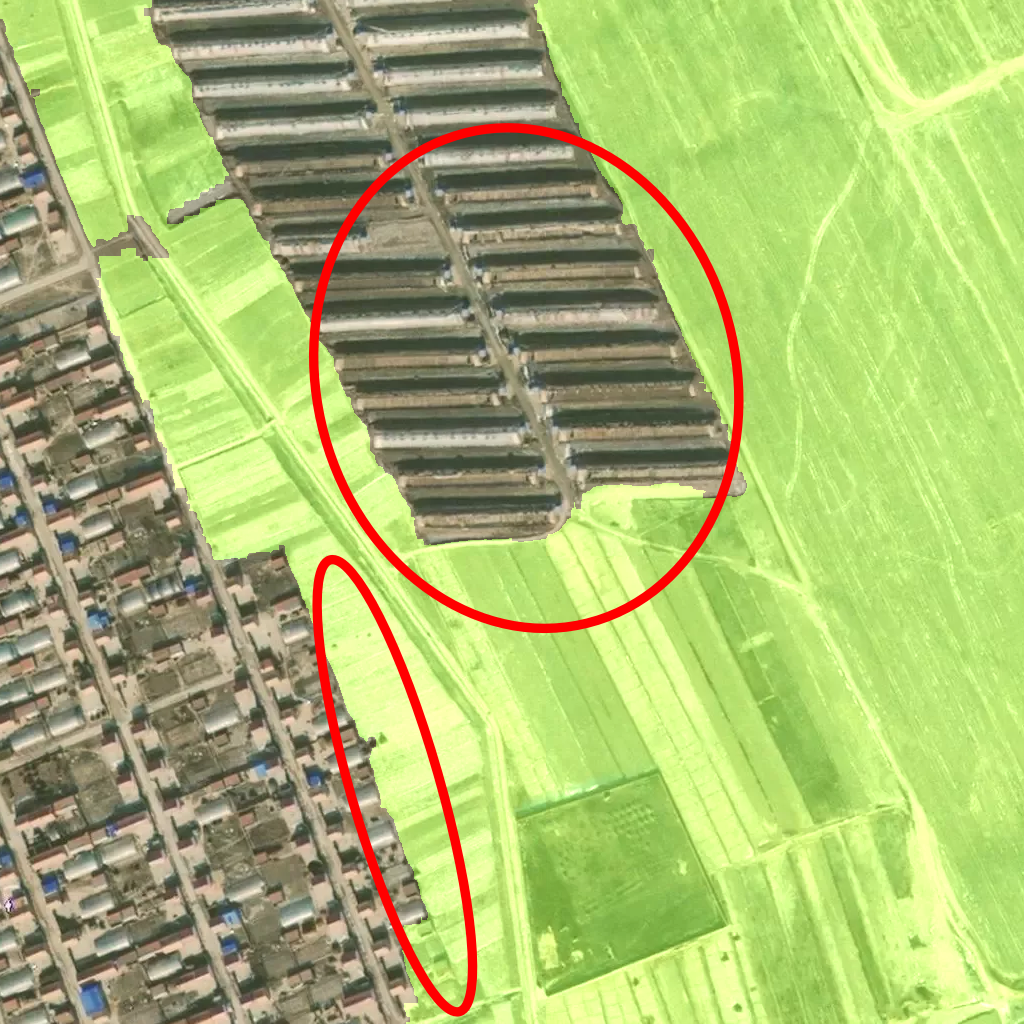}\\
			Image &GT &SL& FT& APT
		\end{tabular}
\vspace{-0.35cm}
\end{table*}
\section{EXPERIMENTS AND RESULTS ANALYSIS}
\subsection{Datasets Description and Experiment Designing}
\subsubsection{Datasets}
In the experiments, we use the GLC product of ESA World Cover\cite{SAM19} as a reference to generate prompts information and validated on two sub-meter-scale cropland datasets from southern and northern China, Which named Hunan and Gansu Datasets. For each datasets, they contains 5000 images with the size of 512×512 from Hunan and Gansu provinces, which includes typical croplands and has RGB (red, green, and blue) bands with a spatial resolution of 0.5m. These images are collected from the Google Earth platform and labeled as cropland and non-cropland.
\subsubsection{Experiment Setup}
We designed supervised learning (SL), fine-tuning (FT), and auto-prompting (APT) experiments. In supervised learning experiments, we utilize 20 labeled sample with the size of 512×512 from target data to directly train the model. In fine-tuning experiments, We take the model trained on other cropland scenes as pretrained model in section \ref{Comparision} and take the vision foundation model from the SAM\cite{SAM13} as pretrained model in section \ref{Discussion}. In auto-prompting experiments, we take the vision foundation model from the SAM as pretrained model and set the positive and negative prompt points number to 30 separately, which feeds it into the model uniformly in three equal parts for iteration. All the above experiments were validated using 5000 labeled samples from the target data
\subsubsection{Training Details}

All experiments were implemented by using PyTorch under the ubuntu 16.04 platform with one GTX3080 (memory 11G). In the fine-tuning experiments, we trained models using the Adam optimizer with a batch size of 4 and 100 epochs. The learning rate was initially set to $1\times 10^{-3}$ and reduced by a factor of $5\times 10^{-4}$. The overall accuracy (OA), Mean Intersection over Union (MIoU) and F1-score are used to compare the performance quantitatively. 

\subsection{Comparison between the traditional approaches and our approach}\label{Comparision}
To investigate the performance of the proposed approach for cropland mapping, we compare our approach with supervised learning (SL) and fine-tuning (FT) approaches on two typical northern and southern cropland datasets (Hunan and Gansu). In baseline approaches, the SL and FT experiments utilized the image encoder architecture from Vision Transformer \cite{SAM13}  configured with the same decoder, and 20 labeled samples with the size of 512×512 used for fine-tuning or directly training. In the proposed auto-prompting approach (APT), we use the prompts generated by the ESA product to guide the model to adapt the target data. 

As shown in Table \ref{tab1_2}, the proposed method presented improved accuracy without any additional manual labeling. Especially in Gansu, the proposed approach gets about 6.67\%  improvement in OA compared with the traditional SL and fine-tune approaches. Although our method only presents about 0.97\% OA enhancement in Hunan region, prompting has a significant optimization in the cropland edges and the completeness of cropland extent, as shown in Table \ref{tab1} with red circles. 

By introducing the “Pretrain+Prompting” paradigm, the proposed approach can obtain a better cropland segment result under the support of GLC Product and visual foundation model. It indicates that the the proposed approach not only achieves better performance without any label cost by using the automatically acquired prompts point to replace the sparsely samples, but also simplified the complex domain gap by guide the model to learning the reasoning ability of each single sample. We will discuss the transfer paradigm and the label efficiency of the auto-prompting with the support of vision foundation model further in the next experiment section. 

\begin{table}[]
	\renewcommand{\arraystretch}{1.45}
	\caption{COMPARISON OF THE TRADITIONAL APPROACHES WITH OUR APPROACH.}
	\label{tab1}
	\resizebox{\hsize}{!}{
	\begin{tabular}{ccccc}
		\hline
		Region & Methods & OA(\%) & Miou(\%) & 		F1-score(\%)    \\ \hline
		\multirow{4}{*}{Hunan} & SL & 77.80 & 62.55 & 76.45 \\
		& FT & 78.72 & 57.78 & 75.98 \\
		& APT(Our method) & \textbf{79.48} & \textbf{62.77} & \textbf{76.57} \\ 
		\hline
		\multirow{4}{*}{Gansu} & SL & 51.86 & 58.95 & 48.92 \\
		& FT & 71.24 & 57.30 & 70.75 \\
		& APT(Our method) & \textbf{75.99} & \textbf{60.72} & \textbf{75.45} \\
		\hline
	\end{tabular}}
\vspace{-0.35cm}
\end{table}

\begin{figure*}[!t]
	\centering
	\subfigure[]{\includegraphics[width=3.5in]{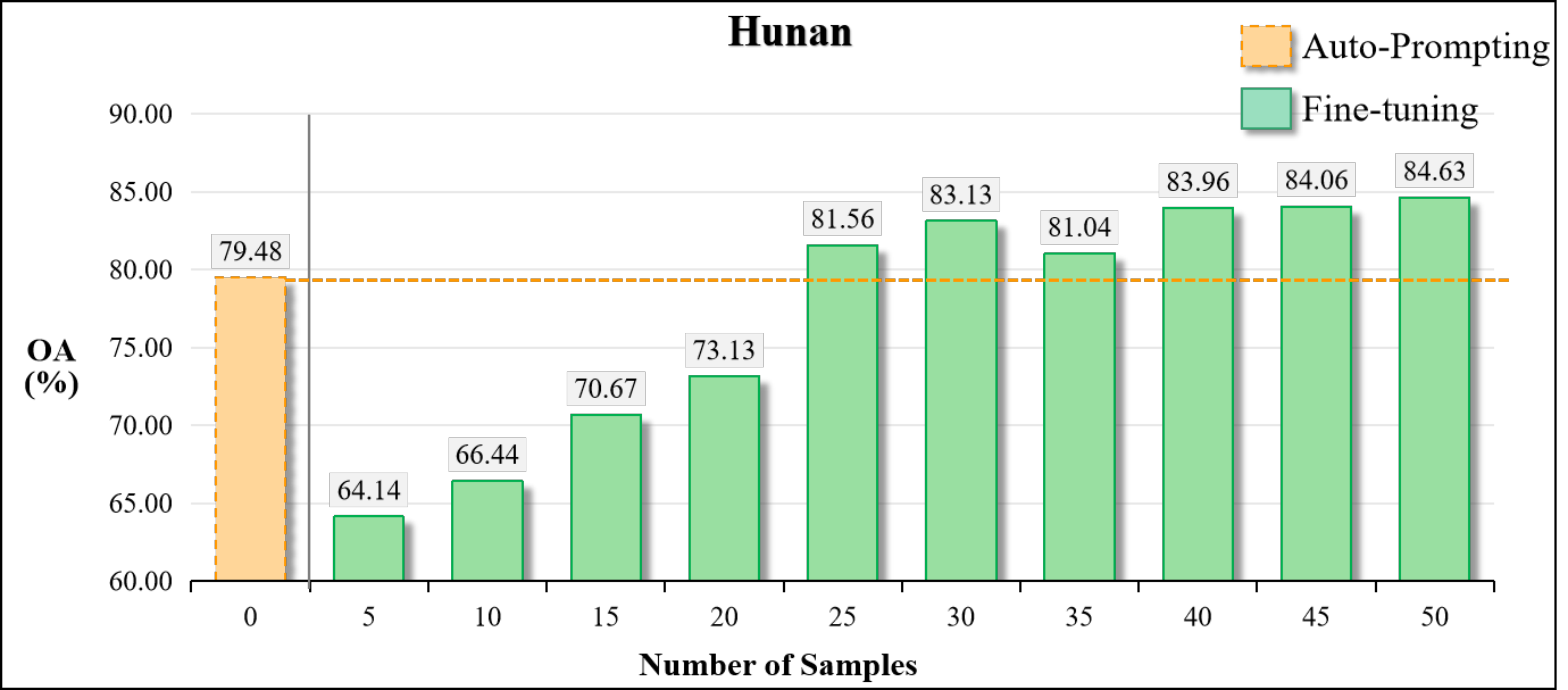}\label{fig:subfig1}}
	\hspace{0.1cm}
	\subfigure[]{\includegraphics[width=3.5in]{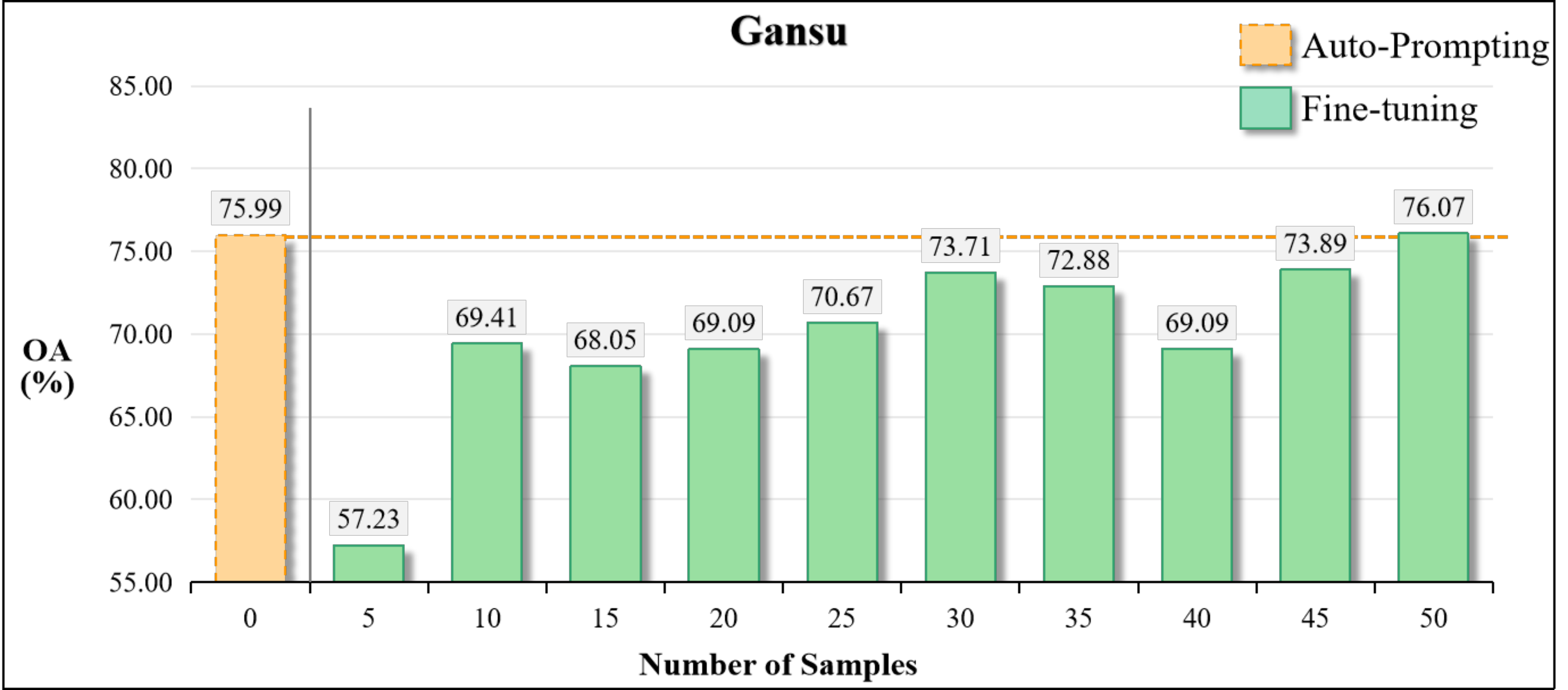}\label{fig:subfig2}}
	\caption{ Comparison of auto-prompting and fine-tuning with different labeled sample size in Hunan(a) and Gansu(b) region}
	\label{fig3}
\vspace{-0.4cm}
\end{figure*}

\subsection{Discussion}\label{Discussion}
In this section, we discuss the importance of the proper way to use the knowledge in pretrained models, and provide a detailed analysis of the benefits in terms of sample labeling efficiency from proposed method. In first part, we compare the “Pretrain+Fine-tuning” and “Pretrain+Prompting” paradigm under the support of vision foundation model from SAM, which demonstrating the importance of using appropriate way to utilize the knowledge embedded in the pre-trained model. In second part,we take different labeled samples size as fine-tuning data compare with the proposed method, which illustrates the labeling cost benefits of using the auto-prompting way to finish the domain adoption process under the support of vision foundation model.

\textbf{Study of the way to transfer:} In this study, We compared the performance of different transfer paradigms supported by vision foundation model (VFM) in the target cropland scenes by using 20 labeled samples. As shown in Table \ref{tab2_1}, In both Hunan and Gansu region, auto-prompting (APT) performs better than fine-tuning (FT) with same pretrained model (improved by nearly 8.69\% and 10.00\% in OA when the target scenes are Hunan and Gansu). It indicates that although the vision foundation model supported by the large amount of data and parameters has a strong potential for feature extraction. Considering the great diversity of cropland features in the target scenes caused by multiple factors (such as topography, climate, aview angle, illumination), it is difficult to represent the complete target scene features distribution by using only a small number of labeled samples in the fine-tuning. With the introduction of prompting, we simplify the difficulty of transfer process by transform the domain adoption between datasets to the domain adoption between datasets and single sample. It reduces the factors that need to be considered during domain adoption process, which improves the generalization ability of the pre-trained model. In the next part, We will analyze the details of how much samples fine-tuning needs to make the vision foundation model adopt the different target cropland scenes. 

\begin{table}[]
	\renewcommand{\arraystretch}{1.45}
	\caption{ACCURACY OF FINE-TUNING AND AUTO-PROMPTING WITH VISION FOUNDATION MODEL.}
	\label{tab2_1}
	\resizebox{\hsize}{!}{
	\begin{tabular}{ccccc}
		\hline
		Region & Transfer Paradigms & OA(\%) & Miou(\%) & F1-score(\%) \\
		\hline
		\multirow{2}{*}{Hunan} & VFM+FT & 73.13 & 61.57 & 72.54 \\
		& VFM+APT & \textbf{79.48} & \textbf{62.77} & \textbf{76.57} \\
		\hline
		\multirow{2}{*}{Gansu} & VFM+FT & 69.09 & 52.55 & 66.02 \\
		& VFM+APT & \textbf{75.99} & \textbf{60.72} & \textbf{75.45} \\
		\hline
	\end{tabular}}
\vspace{-0.5cm}
\end{table}

\textbf{Study of the label efficiency:} In this study, we fine-tuned the vision foundation model using different labeled sample sizes, which explores the sample size is needed under the fine-tuning paradigm to active the potential of the pretrained model. As shown in Figure. \ref{fig3}, when 25 and 50 samples with the size of 512×512 are used, fine-tuning can outperform auto-prompting in Hunan and Gansu region. However, in the process of large-scale high-resolution cropland mapping, pixel-level labeling is undoubtedly manual and time consuming. According to the traditional workflow of cropland mapping, it need to establish amount of localized labeled samples datasets that considering multiple factors such as topography, climate, crop type, planting pattern,view angle, illumination, etc. The proposed auto-prompting method can achieve the delineation of cropland extent anywhere on the global scale without any labeling cost, which can provide technical support for large-scale high-resolution cultivated land mapping.

\section{Conclusion}
In this study, we handle the problem of complex domain gap problem during cropland mapping through prompt learning via the vision foundation model. Moreover, under the “Pretrain+Prompting" paradigm, we utilize the freely available global land cover (GLC) products as the prompt information source to automatically guide the model's reasoning process, which can achieve inexpensive and efficient cropland mapping without any label cost in diverse scenes. By experimentation, the auto-prompting method via the vision foundation model outperforms traditional supervised learning and fine-tuning methods in the case of lacking sufficient labeled data. It provides a possible way to mapping large-scale high-resolution cropland extent without any additional label cost. With the continuous development of the GLC product and vision foundation model, more accurate prompting information and the pre-training models with stronger generalization ability can help the proposed method to obtain better cropland mapping results and even handle more complex scenes. Future work aims at exploring more suitable prompt modes for cropland mapping and investigating the incorporation of expert knowledge in the prompting process.


%





\ifCLASSOPTIONcaptionsoff
  \newpage
\fi



\bibliographystyle{IEEEtran}
\bibliography{bibSAMV2}

\end{document}